\documentclass[10pt,twocolumn,letterpaper]{article}

\usepackage[pagenumbers]{iccv} % To force page numbers, e.g. for an arXiv version

\definecolor{iccvblue}{rgb}{0.21,0.49,0.74}
\usepackage[pagebackref,breaklinks,colorlinks,allcolors=iccvblue]{hyperref}

\usepackage{graphicx}
\usepackage{booktabs}

\usepackage{algorithm}
\usepackage{algorithmic}
\usepackage{bm}
\usepackage{multirow}
\usepackage{float}

\def\paperID{xxx} % *** Enter the Paper ID here
\def\confName{ICCV}
\def\confYear{2025}

\title{Concat-ID: Towards Universal Identity-Preserving Video Synthesis}

\author{Yong Zhong$^1$\thanks{Work done during the internship at Zhipu.} \quad Zhuoyi Yang$^{2}$ \quad Jiayan Teng$^{2}$  \quad Xiaotao Gu$^{3}$ \quad Chongxuan Li$^{1}$\thanks{Correspondence to Chongxuan Li.} \\ \\
$^1$ Gaoling School of AI, Renmin University of China, Beijing, China \\ $^2$ Tsinghua University \quad $^3$ Zhipu AI \\ {\tt\small yongzhong@ruc.edu.cn,\ \{yangzy22,tengjy24\}@mails.tsinghua.edu.cn,} \\ {\tt\small xiaotao.gu@zhipuai.cn,\ chongxuanli@ruc.edu.cn} \\ 
{\textbf{Project page and code:} \url{https://ml-gsai.github.io/Concat-ID-demo/}}
}

\begin{document}
% \maketitle
% \input{sec/0_abstract}    
% \input{sec/1_intro}
% \input{sec/2_formatting}
% \input{sec/3_finalcopy}

% \twocolumn[{%
\maketitle
% \begin{figure}[H]
% \vspace{-8mm}
%         \hsize=\textwidth
%         \centering 
%     \includegraphics[width=1.8\linewidth]{./abstracts/abstrac14.pdf}
%         \caption{\textbf{Concat-ID produces natural videos for identity-preserving video generation.} We select samples for (a) single-identity, (b) multi-identity, and (c) multi-subject scenarios, respectively.}
%         \label{fig:examples}
%         % \vspace{-10pt} 
% \end{figure}
% }]

\begin{abstract}
We present Concat-ID, a unified framework for identity-preserving video generation. Concat-ID employs variational autoencoders to extract image features, which are then concatenated with video latents along the sequence dimension. It relies exclusively on inherent 3D self-attention mechanisms to incorporate them, eliminating the need for additional parameters or modules. A novel cross-video pairing strategy and a multi-stage training regimen are introduced to balance identity consistency and facial editability while enhancing video naturalness. Extensive experiments demonstrate Concat-ID's superiority over existing methods in both single and multi-identity generation, as well as its seamless scalability to multi-subject scenarios, including virtual try-on and background-controllable generation. Concat-ID establishes a new benchmark for identity-preserving video synthesis, providing a versatile and scalable solution for a wide range of applications.
\end{abstract}

% \vspace{-12pt} 

\section{Introduction}
\label{sec:intro}

Identity-preserving video generation, which seeks to create human-centric videos of a specific identity accurately matching a user-provided face image, has recently gained significant attention, as evidenced by the success of commercial tools such as Vidu~\cite{vidu_character2video} and Pika~\cite{pika}. 

A primary challenge in this field is achieving a balance between maintaining identity consistency and enabling facial editability. Prior work \cite{ye2023ip,he2024id,zhang2025magic,huang2025conceptmaster} fails to effectively preserve identity despite utilizing special face encoders and incorporating extra adapters to mitigate cross-modal disparities. To mitigate this limitation, some approaches~\cite{yuan2024identity,fei2025ingredients} substitute the spatially aligned reference image in pre-trained image-to-video models~\cite{blattmann2023stable,yang2024cogvideox} with facial images, leading to a significant improvement in identity consistency. However, they still face challenges in preventing the replication of facial expressions from the reference image. Moreover, the supplementary modules and parameters introduced by these methods contribute to increased complexity in both model training and inference. 

In this work, we introduce Concat-ID, a concise, effective, and versatile framework for identity-preserving video generation. By unifying the model architecture, data processing, and training procedure, Concat-ID not only achieves single-identity video generation but also seamlessly integrates multiple identities and accommodates diverse subjects. Specifically, Concat-ID employs Variational Autoencoders (VAEs) to extract image features, which are then concatenated with video latents along the sequence dimension. This approach relies exclusively on 3D self-attention mechanisms, which are inherently present in state-of-the-art video generation models, to incorporate image features, thereby eliminating the need for extra modules or parameters. Furthermore, to effectively balance identity consistency and facial editability while enhancing video naturalness, we develop a novel cross-video pairing strategy and a multi-stage training regimen.

The quantitative and qualitative results, along with the user study (see~\cref{sec:main_resutls}), demonstrate that Concat-ID produces videos with the most consistent identity and superior facial editability across all baselines, for both single-identity and multi-identity video generation. Moreover, we illustrate that Concat-ID can seamlessly extend to multi-subject scenarios, including virtual try-on and background-controllable generation, while effectively preserving identity (see~\cref{sec:scaling_study}). These findings underscore Concat-ID's capability to scale effectively to diverse subjects, ensuring consistent high performance across various applications.

The principal contributions of this work are as follows:
\begin{itemize}
\item We propose Concat-ID, an effective framework for unified identity-preserving video generation across single-identity, multi-identity, and multi-subject scenarios.
\item Concat-ID utilizes VAEs to extract image features and integrates them via inherent 3D self-attention mechanisms, without introducing additional parameters or modules.
\item We develop a cross-video pairing strategy and a multi-stage training regimen to balance identity consistency and facial editability, while enhancing video naturalness.
\item Concat-ID demonstrates superior identity consistency and facial editability in single and multi-identity scenarios, and seamlessly scales to multi-subject scenarios.
\end{itemize}

% \vspace{-10pt} 

\section{Related works}
\label{sec:related_works}
The rapid advancement of text-to-video and image-to-video diffusion models~\cite{hacohen2024ltx,yang2024cogvideox,kong2024hunyuanvideo,polyak2024movie,zhao2025riflex} has spurred significant interest in fine-tuning these models for downstream tasks, particularly identity-preserving video generation. Tuning-based methods~\cite{ma2024magic,hu2022lora} adapt pre-trained video models for each new identity through test-time fine-tuning. Alternatively, tuning-free methods~\cite{ye2023ip,zhang2025magic,huang2025conceptmaster,he2024id} typically leverage face encoders~\cite{deng2019arcface,radford2021learning} to extract facial features and incorporate additional adapters to mitigate cross-modal discrepancies. Some approaches~\cite{yuan2024identity,wu2024videomaker,fei2025ingredients} further enhance identity consistency by integrating face features extracted from a Variational Autoencoder (VAE). For instance, ConsisID~\cite{yuan2024identity} and Ingredients~\cite{fei2025ingredients} replace spatially aligned reference images in pre-trained image-to-video models for single-identity and multi-identity generation, respectively. Placing greater emphasis on enhancing video naturalness, Movie-Gen~\cite{polyak2024movie} refines the balance between identity consistency and facial editability for single-identity generation through cross-paired data construction. In this work, we explore a unified framework capable of handling single-identity, multi-identity, and multi-subject generation while maintaining a crucial balance between consistency and editability, without requiring test-time fine-tuning.

\section{Preliminary}
\label{sec:preliminary}

Existing state-of-the-art text-to-video and image-to-video models~\cite{hacohen2024ltx,yang2024cogvideox,kong2024hunyuanvideo,polyak2024movie,jin2024pyramidal} generally consist of three main components: a 3D variational autoencoder (VAE) $\mathcal{E}$, text encoders $\mathcal{T}$, and a denoising 
 transformer $\bm\epsilon_{\bm\theta}$. Given a video $\mathbf{X}=\{\mathbf{x}_i\}_{i=1}^{N}$ with $N$ frames,  $\mathcal{E}$ initially compresses the video into a latent representation $\mathbf{Z} \in \mathbb{R}^{T \times HW \times C}$ along the spatiotemporal dimensions, where $HW$ denotes the spatial dimension, $C$ represents the channel dimension, and $T$ is the temporal dimension. To simplify, we refer to $T \times HW$ as the sequence dimension. The $\bm\epsilon_{\bm\theta}$ then takes the noise-corrupted latent representation $\mathbf{Z}$ as its input, and applies a 3D (spatiotemporal) self-attention mechanism~\cite{yang2024cogvideox,hacohen2024ltx} to model the distribution of video content. Additionally, a 3D relative positional encoding (i.e., 3D-ROPE) is incorporated within the 3D attention module to enhance the model’s ability to capture both temporal and spatial dependencies in videos. Meanwhile, the text encoder $\mathcal{T}$ processes the text prompt and encodes it into a text representation $c_{\textrm{txt}}$. $\bm\epsilon_{\bm\theta}$ typically integrates $c_{\textrm{txt}}$ either through cross-attention layers~\cite{polyak2024movie} or by concatenating it with $\mathbf{Z}$~\cite{yang2024cogvideox}. A mean squared error loss~\cite{gao2025diffusionmeetsflow,zhong2024posecrafter} is commonly used to optimize $\bm\epsilon_{\bm\theta}$.

\label{sec:method}

\begin{figure}[tb]
  \centering
  \includegraphics[width=1.0\linewidth]{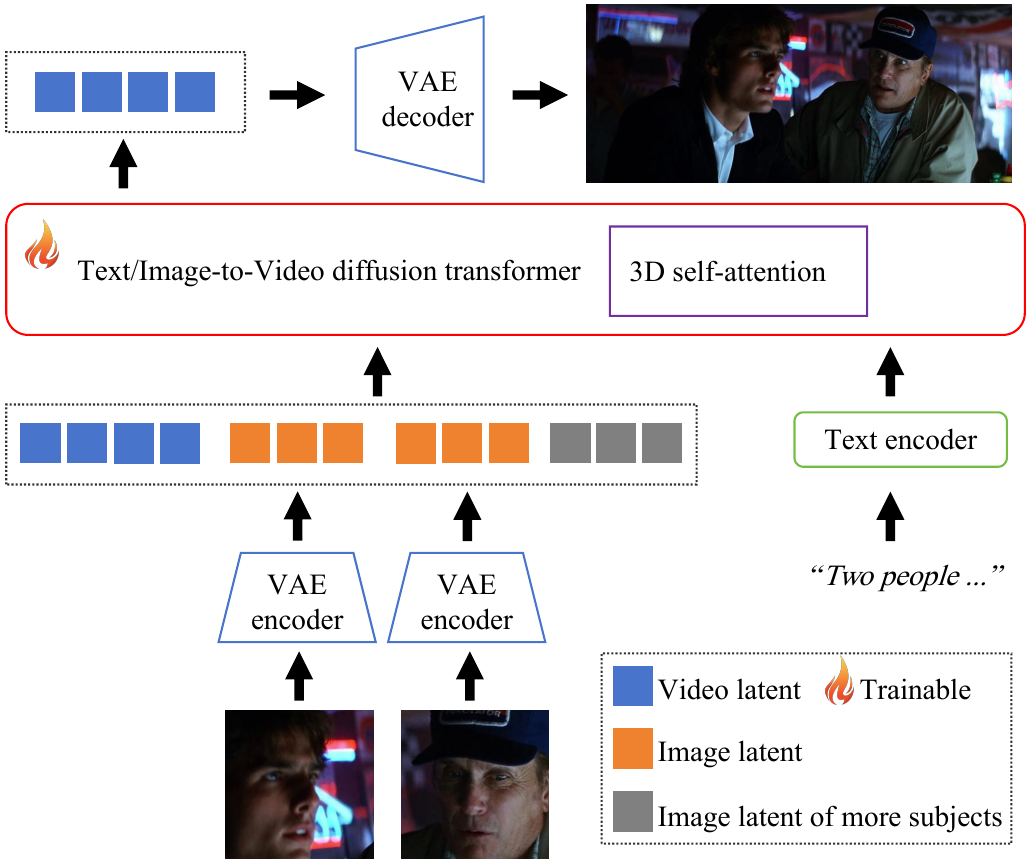}
  \caption{\textbf{The architecture of Concat-ID.} We utilize VAEs to extract image latents from reference images and concatenate them at the end of the video latents along the sequence dimension. Concat-ID relies solely on 3D self-attention mechanisms, which are commonly present in state-of-the-art video generation models, to integrate image features without adding extra modules or parameters.}
  \label{fig:framework}
\vspace{-6pt}
\end{figure}

\section{Concat-ID}
Given a reference image containing a human face, our goal is to generate identity-preserving videos based on user-provided text prompts, while also enabling the integration of additional identities or subjects. To address this challenge, we propose Concat-ID, a concise, effective, and versatile framework. As illustrated in \cref{fig:framework}, we introduce a unified architecture for extracting and injecting features from any number of identities and subjects without requiring extra modules or parameters (see \cref{sec:framework}). To balance identity consistency and facial editability while enhancing video naturalness, we further construct cross-video pairs as training data (see \cref{data_construc}) and propose a novel multi-stage training strategy (see \cref{sec:training_and_inference}).

\subsection{A unified architecture}
\label{sec:framework}

We focus on designing a unified model architecture capable of extracting and fusing the identity feature and readily extendable to multi-identity and multi-subject scenarios. Revisiting the role of VAEs, we recognize their ability to compress conditioning images into the same latent space as the video latent $\mathbf{Z}$. Consequently, our denoising transformer $\bm\epsilon_{\bm\theta}$ can inherently interpret these features. Based on this insight, we adopt the VAE as our feature extractor.

Specifically, for $M$ reference images $\{\mathbf{I}_i\}_{i=1}^M$, we encode each $\mathbf{I}_i$ to obtain the image feature $\mathbf{c}_i = \mathcal{E}(\mathbf{I}_i) \in \mathbb{R}^{1 \times HW \times C}$, and then concatenate these features with $\mathbf{Z}$ in sequence. Thus, the input to $\bm\epsilon_{\bm\theta}$ is given by:
\begin{align}
\label{eq:model_input}
\mathbf{Z'} = \text{Concat}(\mathbf{Z}, \mathbf{c}_1, \mathbf{c}_2, \cdots, \mathbf{c}_M),
\end{align}
where $\text{Concat}(\cdot, \cdot, \cdots)$ denotes concatenation along the sequence dimension and $\mathbf{Z'} \in \mathbb{R}^{(N+M) \times HW \times C}$. As shown in \cref{fig:framework}, this feature injection through concatenation is compatible with any video generation model that utilizes 3D self-attention, which are generally present in state-of-the-art video generation models. Since $\mathbf{Z}$ and $\mathbf{c}_i$ are in the same latent space, $\bm\epsilon_{\bm\theta}$ can seamlessly integrate identity-preserving features without the need for additional modules or parameters to address cross-modal disparities.

Concatenating $\mathbf{Z}$ and $\mathbf{c}_i$ along the channel dimension is another direct method for feature injection, as employed in ConsisID~\cite{yuan2024identity} and Ingredients~\cite{fei2025ingredients}. However, this strategy introduces artifacts (see~\cref{fig:qualitative_comparison} and~\cref{fig:qualitative_comparison_2_people}) due to spatial misalignment between face images and video latents. In contrast, by leveraging a 3D self-attention mechanism, our sequence concatenation promotes spatial interactions without compromising the quality of any generated frame. Furthermore, it scales efficiently to handle multi-identity and multi-subject scenarios (see~\cref{fig:examples}).

\subsection{Data construction}
\label{data_construc}
\begin{figure}[tb]
  \centering
\begin{subfigure}{\linewidth}
  \centering
        \includegraphics[width=1.0\linewidth]{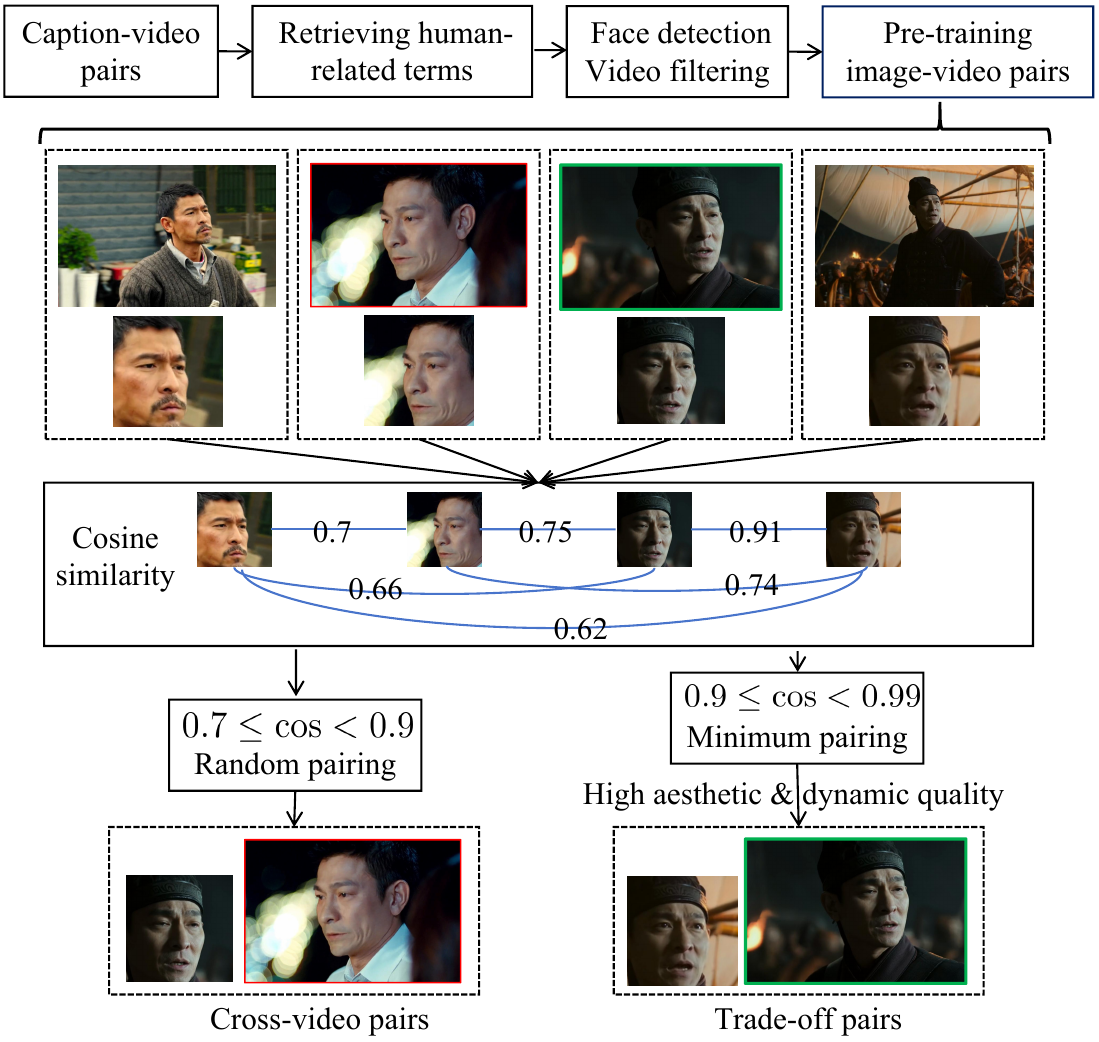}
        \caption{The procedure of data processing.}
    \end{subfigure}
  \centering
\begin{subfigure}{\linewidth}
  \centering
        \includegraphics[width=1.0\linewidth]{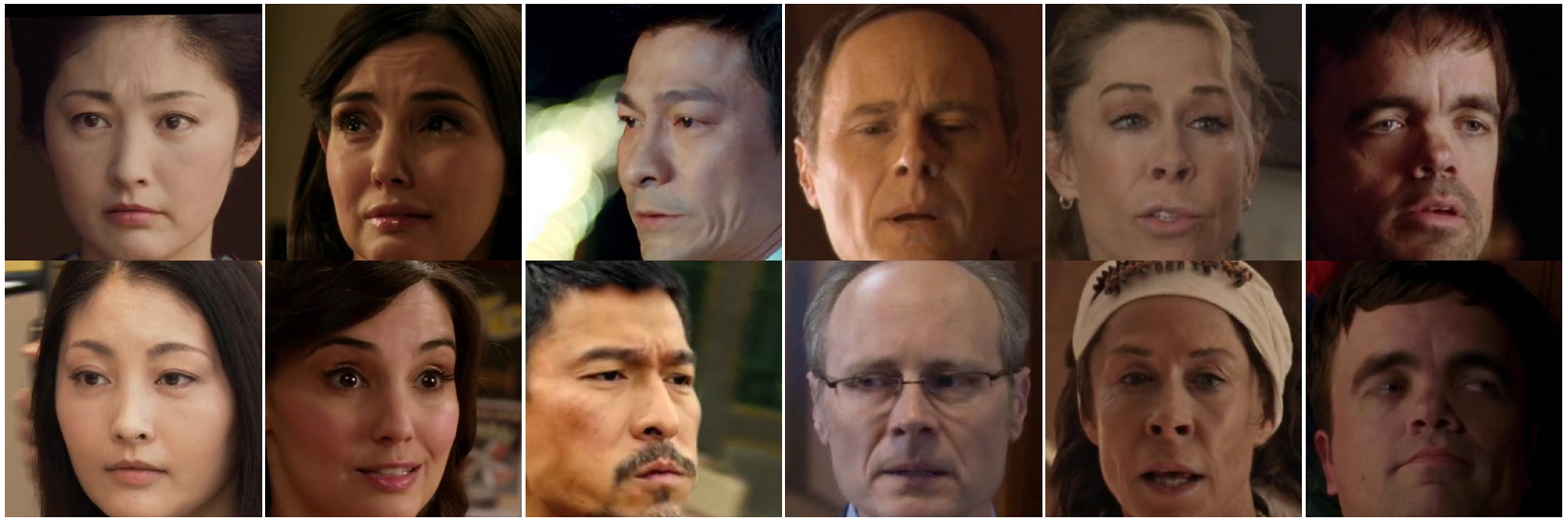}
        \caption{Some samples of paired cross-video reference images.}
  \label{fig:cross_video}
    \end{subfigure}
\begin{subfigure}{\linewidth}
  \centering
        \includegraphics[width=1.0\linewidth]{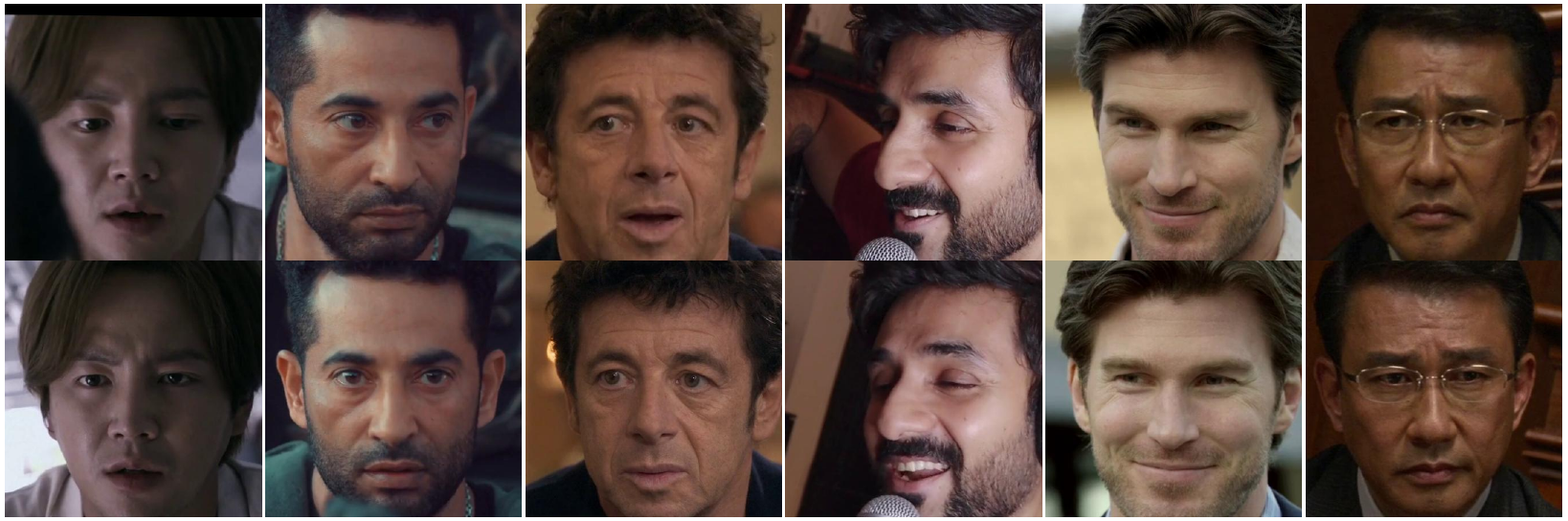}
        \caption{Some samples of trade-off pairs.}
  \label{fig:trade-off}
    \end{subfigure}
  \caption{\textbf{Constructing three types of image-video pairs for a single identity:} pre-training, cross-video and trade-off pairs.}
  \label{fig:data_pipeline}
\vspace{-1pt} 
\end{figure}

The task of identity-preserving video generation relies on image-video pairs as training data, where an image must depict a human face that matches the identity of corresponding videos. To progressively balance identity consistency and facial editability, as illustrated in \cref{fig:data_pipeline}, we construct three types of image-video pairs for a single identity: pre-training pairs $\mathcal{S}_{\text{pre}}$, cross-video pairs $\mathcal{S}_{\text{cross}}$, and trade-off pairs $\mathcal{S}_{\text{trade}}$. 

\noindent
\textbf{Pre-training pairs.} To ensure data quality, we filter out videos that are unrelated to humans, contain inconsistent numbers of individuals, or exhibit inconsistencies in identity. Specifically, to retrieve human-related videos from the caption-video pairs, we design a human term table that includes various categories such as basic human descriptors, gender, and occupation. We then exclude videos whose captions do not contain any human-related terms. Next, we uniformly sample two frames per second from each video, detect faces using SCRFD~\cite{guo2021sample}, and remove videos if more than 30\% of the frames have inconsistent numbers of individuals.\footnote{The common person count across frames is considered the video's person count.} Finally, for frames with the same face count, we compute the ArcFace cosine similarity~\cite{deng2019arcface} between consecutive frames and discard videos if more than 30\% of the frames have a similarity score below 0.5.

The above processes yield 1.3 million videos featuring a single identity, and we uniformly select 5 face images per video, defining pre-training pairs $\mathcal{S}_\textrm{pre}=\{(\mathbf{I}^{k}_i,\mathbf{X}^{k})\}_{i,k}$ where $k$ denotes the video index and $\mathbf{I}^{k}_i$ represents the $i$-th reference image of $\mathbf{X}^{k}$. The self-supervised nature of this paired data, where images from the same video serve as labels, inherently limits facial editability. Specifically, models trained on such data may produce frames in which facial expressions unintentionally mirror those of the reference images (see~\cref{fig:ablation}), leading to unnatural content. This issue becomes particularly pronounced when a semantic gap exists between the reference images and the text prompts. To enhance facial editability and naturalness, we propose a cross-video image-video pairing strategy.

\noindent
\textbf{Cross-video pairs.} The standard process for constructing video clips involves segmenting raw long videos into multiple shorter segments using various algorithms that detect scene transitions, such as motion variations and shot changes. Theoretically, many existing video clips in training sets feature varied facial expressions and head poses of the same person. To construct cross-video pairs where the reference image originates from a different video, we calculate the cosine similarity among images $\{\mathbf{I}^{v}_1\}_{v}$. For the $k$-th video, we randomly select an image $\mathbf{I}^j_1$ from $\{\mathbf{I}^{v}_1\}_{v}$ as its paired reference image, ensuring that $0.7 \leq \cos(\mathbf{I}^j_1, \mathbf{I}^k_1) < 0.9$, where the function $\cos(\cdot, \cdot)$ computes the cosine similarity. The final cross-video pairs $\mathcal{S}_{\text{cross}}$ include 0.8 million image-video pairs with 0.5 million reference images, indicating a reference image can correspond to multiple videos.

Personalized image generation can also synthesize reference images with the same identity as given videos but varied identity-irrelevant factors, as demonstrated in~\cite{he2024imagine,polyak2024movie}. However, this approach incurs high computational costs, particularly for large-scale image-video pairs. Additionally, existing personalized generation methods~\cite{ye2023ip,guo2024pulid,zhou2024toffee} often struggle to preserve detailed facial features, which limits their effectiveness. In contrast, as shown in \cref{fig:cross_video}, our retrieval-based method efficiently gathers a large-scale set of real reference images that accurately match the identity of corresponding videos while exhibiting diversity across multiple dimensions, such as facial expressions, hairstyles, lighting conditions, and other identity-irrelevant factors. 

\noindent
\textbf{Trade-off pairs.} Similar to the construction of cross-video pairs, for the $k$-th video, we identify its reference image $\mathbf{I}^j_1$ with the smallest $\cos(\mathbf{I}^j_1, \mathbf{I}^k_1)$, ensuring that $0.9 \leq \cos(\mathbf{I}^j_1, \mathbf{I}^k_1) < 0.99$. This forms our trade-off dataset $\mathcal{S}_\textrm{trade}$ with 160 thousand videos, improving consistency between reference images and videos compared to cross-video pairs. Additionally, we filter out videos where the facial region occupies less than 4\% or more than 90\% of the frame area and rank $\mathcal{S}_\textrm{trade}$ based on the weighted sum of aesthetics scores, optical flow scores, and motion scores~\cite{yang2024cogvideox}. Finally, we retain the top 50,000 videos for training. 

In this section, we detail the data construction process for a single identity. However, this procedure can be seamlessly scaled to multi-identity by independently processing each identity within a video. Similarly, it can be extended to general subjects by replacing face detectors with open-set detectors, such as Grounding DINO~\cite{ren2024grounding}, and substituting ArcFace cosine similarity with general feature similarity metrics, such as CLIP cosine similarity~\cite{radford2021learning}. Please refer to Appendix B for further details on the training data construction for our multi-identity and multi-subject scenarios. 

\subsection{Training strategy}
\label{sec:training_and_inference}

Building on our innovative data construction, we introduce a multi-stage training process: pre-training stage, cross-video fine-tuning, and trade-off fine-tuning.  In the pre-training stage, we optimize a text-to-video model on $\mathcal{S}_\textrm{pre}$ to map facial details into generated videos. This self-supervised training method may constrain certain generated video frames to adhere strictly to the given condition images, potentially degrading the editability of facial expressions and the overall naturalness. The cross-video fine-tuning on $\mathcal{S}_\textrm{cross}$, using image-video pairs derived from different videos, can alleviate this issue. However, we observe that this fine-tuning enhances facial editability at the expense of identity fidelity (see~\cref{sec:ablation}). 

A simple strategy to further balance fidelity and editability is to mix pre-trained pairs and cross-video pairs in a 1:1 ratio, a similar method adopted by Movie-Gen~\cite{polyak2024movie}. However, our initial experiments suggest that this approach results in unstable training due to varying identity consistency between pre-trained pairs and cross-video pairs. To address this issue while ensuring high-degree motion and high artistic quality, we ultimately fine-tune the model on $\mathcal{S}_\textrm{trade}$.

Throughout all training stages, we proportionally scale, pad, and center-crop images to match the video resolution. To ensure the model focuses on facial regions during training and prevents background leakage during inference, we segment and drop the background of reference images~\cite{xie2021segformer}. Additionally, to improve robustness and generalization, we introduce random noise to reference images during training, while omitting this noise during inference. To further differentiate the image latent $\mathbf{c}_i$ from the video latent $\mathbf{Z}$ and distinguish between different $\mathbf{c}_i$, we extend 3D-RoPE to incorporate multiple reference images along the sequence dimension. Specifically, we introduce a temporal bias $N$ to define the 3D position of a token $\mathbf{t}_{h,w}$ in $\mathbf{c}_i$:
\begin{align}
\text{3D-Pos}(\mathbf{t}_{h,w}) = (i + N, h, w),
\end{align}
where $\text{3D-Pos}(\cdot)$ denotes the 3D position and $(h,w)$ are the spatial coordinates of the token.

Owing to the simplicity and efficiency of Concat-ID in both data construction and model architecture, our training strategy can seamlessly scale to multi-identity and multi-subject scenarios. Moreover, we establish that single-identity pre-training facilitates enhanced identity preservation in these downstream tasks (see~\cref{tab:quantitative_abalation_multi-identity}).

\section{Experiments}
\label{sec:experiments}

\subsection{Experimental settings}
\label{sec:settings}

\noindent
\textbf{Datasets.} We evaluate all methods on the ConsistID-Benchmark~\cite{yuan2024identity}, which consists of 172 reference images and 90 text prompts spanning nine categories. To ensure a fair comparison, we exclude reference images present in our training data using a combination of automated and manual filtering techniques. Consequently, our evaluation dataset comprises 873 prompt-image pairs, derived from 97 reference images, with one prompt randomly selected from each category for each image. For multi-identity evaluation, we additionally construct 14 distinct pairs of reference images and design 20 textual prompts using ChatGPT~\cite{achiam2023gpt}. Please refer to Appendix A.1 for further details.

\noindent
\textbf{Metrics.} We evaluate all methods on identity consistency, text alignment, and facial editability. (1) Identity consistency: Following \cite{yuan2024identity}, we use FaceSim-Arc (ArcSim) and FaceSim-Cur (CurSim) to assess the average cosine similarity between reference images and generated videos based on ArcFace~\cite{deng2019arcface} and CurricularFace~\cite{huang2020curricularface}, respectively. These face recognition models are specifically designed to disentangle identity-related features from identity-unrelated ones. (2) Text alignment: We adopt ViCLIP~\cite{wang2023internvid} to compute the similarity between text prompts and generated videos, following \cite{huang2024vbench, polyak2024movie}. (3) Facial editability: We calculate the cosine distance of CLIP image embeddings~\cite{radford2021learning} (CLIPDist) between reference images and video frames. CLIP effectively captures comprehensive facial features, and thus a larger CLIPDist indicates improved facial editability.

\noindent
\textbf{Implementation details.} We use the text-to-video model CogVideoX-5B~\cite{yang2024cogvideox} as our base model. The learning rates are set to $1.0 \times 10^{-5}$, $5.0 \times 10^{-6}$, and $5.0 \times 10^{-6}$ for the first, second, and third training stages, respectively. We fine-tune all model parameters with a linear learning rate decay across all stages. The training data resolution is maintained at $480 \times 720$ pixels with 49 frames per video. Text and image prompts are independently dropped with a probability of 0.1. Further details are provided in Appendix A.2.

\noindent
\textbf{Baselines.} For a comprehensive comparison, we use three representative open-source approaches as baselines. (1) Single-identity personalization methods: ID-Animator~\cite{he2024id} and ConsisID~\cite{yuan2024identity}. (2) Multi-identity personalization methods: Ingredients~\cite{fei2025ingredients}. ID-Animator, ConsisID, and Ingredients all incorporate additional adapters and auxiliary loss functions to enhance identity consistency. Notably, Concat-ID, ConsisID, and Ingredients are all built upon the same video model, CogVideoX-5B.

\subsection{Main results}
\label{sec:main_resutls}
We demonstrate the effectiveness of Concat-ID through quantitative metrics, qualitative assessments, and the user study for single-identity and multi-identity generation.

\noindent
\textbf{Quantitative comparisons.} \Cref{tab:quantitative_comparision} presents the quantitative results for single-identity and multi-identity generation. 
For single-identity generation, ID-Animator performs the worst, exhibiting the lowest ArcSim, CurSim, and CLIPDist scores. This suggests that it achieves the least effective balance between identity preservation and facial editability. Moreover, ID-Animator, ConsisID, and Ingredients incorporate additional adapters and auxiliary loss functions to enhance identity consistency, increasing the complexity of both training and generation processes.

In contrast, for both single-identity and multi-identity generation, Concat-ID achieves superior identity consistency simply by concatenating image latents after video latents, highlighting the effectiveness of our architecture. Furthermore, by constructing cross-video pairs, Concat-ID attains a higher CLIPDist score than ID-Animator, ConsisID, and Ingredients, demonstrating an optimal balance between identity preservation and facial editability.

\begin{table}[tb]
\setlength{\tabcolsep}{5pt}
  \centering
\resizebox{\linewidth}{!}{
  \begin{tabular}{lccccc}
    \toprule
\multirow{2}[1]{*}{Method}  & \multicolumn{2}{c}{\textbf{Identity consistency}} & \multicolumn{1}{c}{\textbf{Text alignment}} & \multicolumn{1}{c}{\textbf{Facial editability}} \\ 
\cmidrule(lr){2-3} \cmidrule(lr){4-4} \cmidrule(lr){5-5}
 &  ArcSim $\uparrow$ & CurSim $\uparrow$ & ViCLIP $\uparrow$ & CLIPDist $\uparrow$ \\
 \midrule
 \multicolumn{5}{c}{Single identity} \\
    \midrule
    ID-Animator~\cite{he2024id} $^{\ddagger}$& 0.289 & 0.304 & 0.204 & 0.297\\
    ConsisID~\cite{yuan2024identity}$^{\dagger\ddagger}$ & 0.432 & 0.451 & 0.237 & 0.303\\
    Concat-ID (Ours) $^{\dagger}$ & \bf{0.442}& \bf{0.466} & \bf{0.242} & \bf{0.325} \\
 \midrule
 \multicolumn{5}{c}{Multiple identities} \\
    \midrule
    Ingredients~\cite{fei2025ingredients} $^{\dagger\ddagger}$ & 0.293 & 0.316  &  \bf{0.199} & 0.407\\
    Concat-ID (Ours) $^{\dagger}$ & \bf{0.492}& \bf{0.514} & 0.190 & \bf{0.410} \\
    \bottomrule
  \end{tabular}
}
  \caption{
  \textbf{Quantitative results for single-identity and multi-identity generation.} ${\dagger}$ denotes that these methods share the same video model. $\ddagger$ indicates corresponding methods introduce additional adapters and auxiliary loss. Concat-ID achieves superior identity consistency and facial editability while maintaining better or comparable text alignment relative to the baselines. 
} 
  \label{tab:quantitative_comparision}
% \vspace{-10pt}
\end{table}

\begin{figure*}[tb]
  \centering
  \includegraphics[width=0.8 \linewidth]{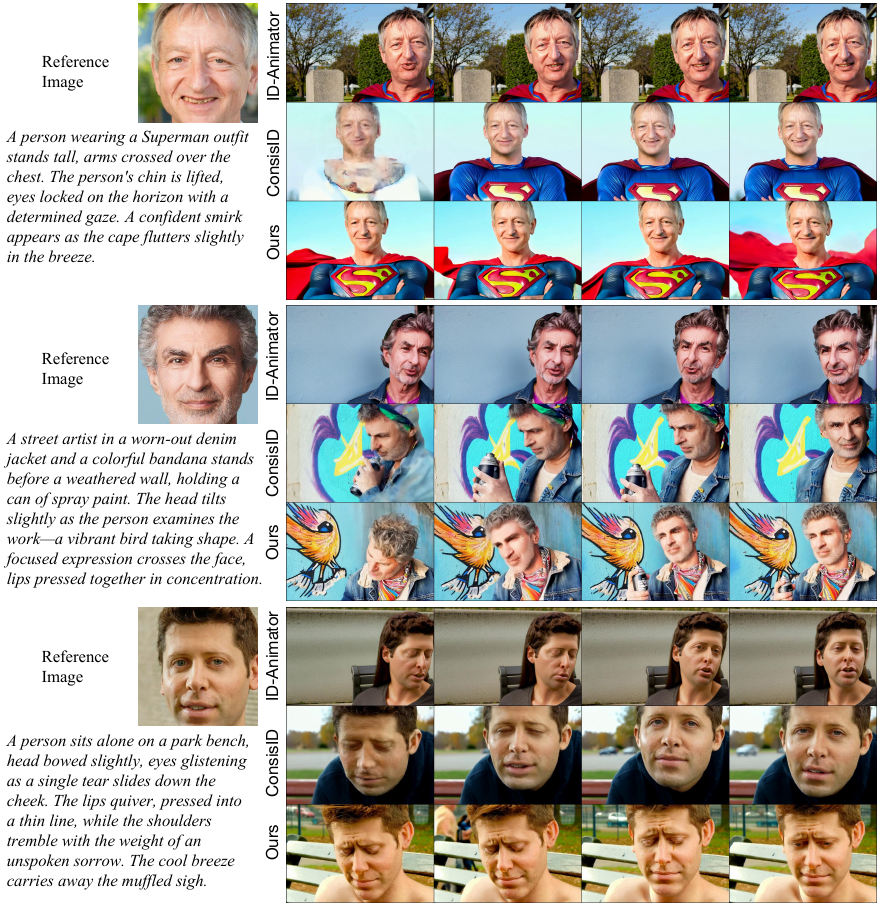}
  \caption{\textbf{Qualitative comparisons for single-identity generation.} ID-Animator fails to preserve facial details, while ConsisID replicates the expressions of the reference images, particularly in the third case, where the semantic gap between texts and reference is significant. Concat-ID effectively preserves identity, while simultaneously preventing the direct replication of facial expressions from reference images.}
  \label{fig:qualitative_comparison}
\vspace{-10pt} 
\end{figure*}

\begin{figure}[tb]
  \centering
  \includegraphics[width=1.0\linewidth]{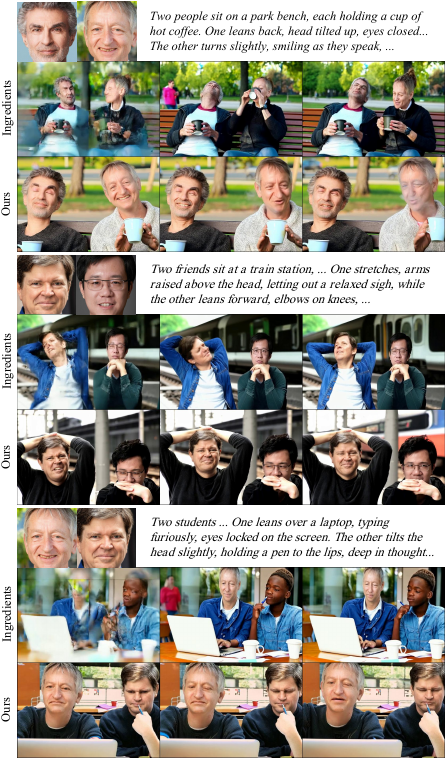}
  \caption{\textbf{Qualitative comparisons for multi-identity generation.} Concat-ID better maintains different identities.}
  \label{fig:qualitative_comparison_2_people}
% \vspace{-10pt} 
\end{figure}

\noindent
\textbf{Qualitative comparisons.} ~\cref{fig:qualitative_comparison} presents qualitative comparisons for single-identity generation. ID-Animator fails to maintain facial characteristics. ConsisID achieves better identity consistency, but some frames replicate facial expressions of reference images. In contrast, Concat-ID mitigates this issue while preserving identity by leveraging advantages of cross-video pairs. For multi-identity generation, as shown in~\cref{fig:qualitative_comparison_2_people}, Concat-ID produces videos that more accurately match identities in given images compared to Ingredients, demonstrating its effectiveness and scalability. 

To maximize the potential of image-to-video models, ConsisID and Ingredients concatenate the reference image with the first latent frame along the channel dimension. However, this feature injection approach can introduce artifacts in the first generated frame due to spatial misalignment between faces images and generated videos, as evident in the initial frames of all videos. As a comparison, Concat-ID excels in identity preservation without compromising the quality of any generated frames, highlighting the validity of our concatenation along the sequence dimension.

\begin{figure}[tb]
  \centering
  \includegraphics[width=1.0 \linewidth]{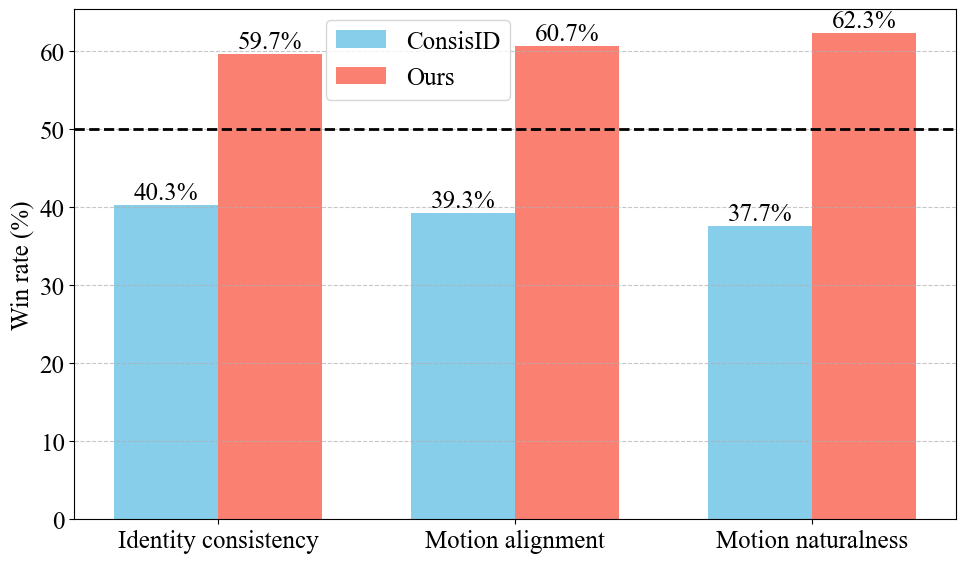}
  \caption{\textbf{Human evaluation.} Concat-ID produces more precise and natural videos while effectively preserving identity.}
  \label{fig:user_study}
% \vspace{-9pt} 
\end{figure}

\noindent
\textbf{User study.} According to both quantitative and qualitative results, we compare Concat-ID with the strongest baseline, ConsisID, through human evaluation. Specifically, we generate 100 videos using 10 reference images and 10 prompts designed by ChatGPT~\cite{achiam2023gpt} to focus on expression and head pose variation. For each video group, voters answer three questions, selecting the video that: (1) best matches the reference image in facial similarity (identity consistency), (2) best aligns with the facial expressions and head poses described in the prompt (facial motion alignment), and (3) exhibits the most natural and smooth facial motion (facial motion naturalness). With 100 video groups, three types of questions, and three voters participating, we collect a total of 900 video comparison results. As shown in \cref{fig:user_study}, Concat-ID surpasses ConsisID by a significant margin in identity consistency and motion alignment and naturalness, demonstrating the effectiveness of our architecture and the advantages of cross-video pair construction.

\subsection{Multiple identities and subjects}
\label{sec:scaling_study}

We demonstrate that the architecture, data construction, and training strategy of Concat-ID make it seamlessly extendable to multi-identity and multi-subject scenarios.

\noindent
\textbf{Multi-identity scenarios.} As illustrated in \cref{fig:examples}b, when provided with face images of different individuals, Concat-ID can generate multi-person videos while preserving their identities, without requiring any additional parameters or modules compared to single-identity generation. Notably, despite being trained on only 40,000 videos, Concat-ID can generate three-identity videos while maintaining distinct identities, leveraging the prior knowledge from two-identity pre-training and a powerful 3D self-attention mechanism that effectively captures both temporal and spatial dependencies. Moreover, Concat-ID determines the spatial position of each identity in the generated videos based on the concatenation sequence of the reference images. 

\noindent
\textbf{Multi-subject scenarios.} As illustrated in \cref{fig:examples}c, by sequentially concatenating a face image with a clothing image, Concat-ID enables virtual try-on while preserving both the given identity and intricate clothing details, such as logos and textures. This capability also highlights Concat-ID's potential in simulating interactions between people and objects. Furthermore, the background-controllable identity-preserving generation achieved by Concat-ID demonstrates its ability to manipulate spatial layouts in generated videos by integrating spatially aligned conditions.

In this section, we introduce two-identity and three-identity generation, along with two additional subjects (\ie clothing and background). Further details on training and data are provided in Appendix B. We posit that Concat-ID's architecture, characterized by its simplicity and effectiveness, coupled with the generalizability of its data construction and training strategy, enables effective scalability to more identities and diverse subjects, ensuring consistent high performance across a wider range of applications.

\subsection{Ablation study}
\label{sec:ablation}
\begin{figure}[tb]
  \centering
  \includegraphics[width=1.0\linewidth]{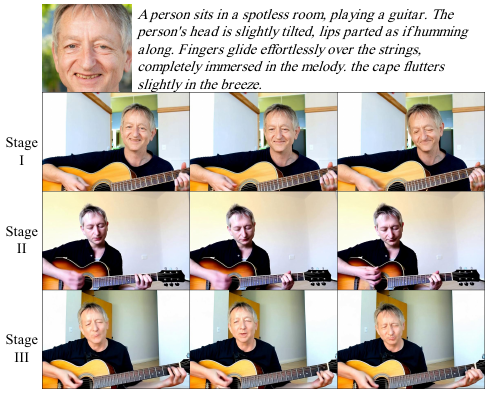}
  \caption{\textbf{Qualitative ablation.} Stage I, Stage II, and Stage III indicate the pre-training stage, cross-video stage, and trade-off stage.}
  \label{fig:ablation}
% \vspace{-10pt} 
\end{figure}

\cref{fig:ablation} present the qualitative ablation of Concat-ID. The pre-training stage achieves the best identity consistency but results in low facial editability. For example, facial expressions of some frames in the pre-training stage closely resemble those in reference images. However, the cross-video stage enhances editability at the expense of identity consistency, aligning with the findings in \cite{polyak2024movie}. In the third stage, Concat-ID further refines the matching threshold of cross-video pairs to better balance identity preservation and facial editability. Leveraging prior knowledge from both pre-training and cross-video fine-tuning, the trade-off stage achieves an optimal balance using only 50,000 videos. These results underscore the effectiveness of each stage in our training strategy. Moreover, the quantitative analysis in Appendix C consistently supports our findings.

We also investigate the influence of single-identity pre-training on multi-identity and multi-subject pre-training. Specifically, we conduct a comparative analysis of Concat-ID with and without single-identity pre-training. Although the two-identity generation is pre-trained on approximately 0.3 million videos, as presented in \cref{tab:quantitative_abalation_multi-identity}, single-identity pre-training still results in improved ArcSim and CurSim scores across all identities. This enhancement indicates that single-identity pre-training effectively strengthens identity preservation in downstream tasks. These findings provide empirical support for the scalability of our architecture, data construction methodology, and training strategy.

\begin{table}[tb]
  \centering
\resizebox{\linewidth}{!}{
  \begin{tabular}{lccccc}
    \toprule
\multirow{2}[1]{*}{Method}  & \multicolumn{2}{c}{\textbf{Identity-1}} & \multicolumn{2}{c}{\textbf{Identity-2}}  \\ 
\cmidrule(lr){2-3} \cmidrule(lr){4-5} 
 &  ArcSim $\uparrow$ & CurSim $\uparrow$ &  ArcSim $\uparrow$ & CurSim $\uparrow$\\
    \midrule
    No single-identity pre-training & 0.514 & 0.535 & 0.526 & 0.550\\
    Concat-ID (Pre-training) & \bf{0.629} & \bf{0.650} & \bf{0.651} & \bf{0.674}\\
    \bottomrule
  \end{tabular}
}
  \caption{\textbf{The effect of single-identity pre-training on multi-identity pre-training.} The single-identity pre-training enhances identity consistency in downstream tasks.}
  \label{tab:quantitative_abalation_multi-identity}
% \vspace{-10pt}
\end{table}

\section{Conclusions}
In this paper, we introduce Concat-ID, a unified framework for identity-preserving video generation. Concat-ID relies solely on 3D self-attention mechanisms, which are commonly used in state-of-the-art video generation models, without introducing additional modules or parameters. We also present a novel cross-video pairing strategy and a multi-stage training regimen to balance identity consistency and facial editability while enhancing video naturalness. Thanks to its architecture, data construction, and training strategy, Concat-ID can scale seamlessly to multi-identity and multi-subject scenarios.

\noindent
\textbf{Limitations.} Similar to common video generation models, our approach faces challenges in preserving the integrity of human body structures, such as the number of fingers, when handling particularly complex motions.
In this paper, we focus on the single-identity scenario, and further improvement and evaluation  of Concat-ID's performance in multiple-identity and multi-subject scenarios is left for future work.

\clearpage
{
    \small
    \bibliographystyle{ieeenat_fullname}
    \bibliography{main}
}

\clearpage

\appendix

\onecolumn

\begin{center}
\textbf{\Large Supplementary material}
\end{center}

\section{Experimental settings}
\label{sec:experiments_settings}

\subsection{Datasets}
We remove reference images from the ConsistID-Benchmark that may appear in our training data using both manual and automated filtering methods. (1) Manual filtering: For each reference image in the ConsistID-Benchmark, we compute its cosine similarity with all training images and identify the most similar one. Human evaluators then determine whether the two images depict the same person. If so, all reference images of the corresponding identity are excluded. (2) Automated filtering: All reference images of an identity are discarded if any training image has a cosine similarity greater than 0.45 with one of its reference images.

\subsection{Implementation details}
In the first stage, we randomly select one reference image from a set of five for each video. Traditional data augmentation techniques, such as flipping, are not used for face images, as they can cause data augmentation leakage~\cite{karras2020training}, leading the model to learn the augmented data distribution rather than the original distribution. For instance, horizontal flipping may result in incorrectly mirrored faces in generated videos. 

\subsection{Baselines} 
We try our best not to change original settings of baselines to maintain their original capabilities. IDAnimator~\cite{he2024id} and ConsisID~\cite{yuan2024identity} can produce 16-frame and 49-frame videos at a resolution of $480\times720$, respectively. The multi-identity baseline Ingredients~\cite{fei2025ingredients} generates 49-frame videos at a resolution of $480\times720$, integrating two distinct identities.

\subsection{Training cost} 
The first, second, and third stages of training in the single-identity scenario required 3,260, 2,104, and 135 NVIDIA H800 GPU hours, respectively, with the cost of the third stage being negligible.

\section{Multiple identities and subjects}

\subsection{Multi-identity scenarios}
Through the data construction process of pre-training pairs, we obtain approximately 300,000 videos featuring two identities. For each identity, we determine the sequence order by computing the mean horizontal position of face boxes across all reference images. We discard reference images where the face position does not align with the determined sequence order. Next, we construct cross-video pairs by independently processing each identity within a video. Finally, we collect around 8,000 videos, each of which contains identities that have corresponding cross-video reference images.

A similar strategy is used to construct three-identity training data, resulting in a final dataset of approximately 40,000 pre-training videos. For cross-video pairs, we retain videos in which at least two identities have corresponding cross-video reference images, resulting in about 2,000 videos.

For multiple identities, the pairing cosine similarity ranges between 0.87 and 0.97. We initialize the model using single-identity pre-training weights and train it only on the first two stages (i.e., the pre-training stage and cross-pair fine-tuning stage). Our findings indicate that single-identity pre-training facilitates multi-identity convergence and enhances identity consistency.

\subsection{Multi-subject scenarios}
We select a subset from the cross-video pairs of single-identity, comprising approximately 200,000 videos, where the pairing cosine similarity ranges between 0.87 and 0.97. To achieve virtual try-on, we use Grounded-SAM-2\footnote{\url{https://github.com/IDEA-Research/Grounded-SAM-2}} to detect and segment the clothing of identities. For background-controllable generation, we extract the first frame and use Grounded-SAM-2 to obtain human masks. We then apply SDXL\footnote{\url{https://huggingface.co/diffusers/stable-diffusion-xl-1.0-inpainting-0.1}} to inpaint the masked areas to get bacground images, using a randomly selected classification label from YOLO\footnote{\url{https://github.com/ultralytics/ultralytics}} as input prompts.

We use weights from single-identity pre-training as initialization and apply only random horizontal flip augmentation to clothing images. Additionally, we introduce random noise to both the background and clothing images during training. In multi-subject scenarios, we only train models on the cross-pair fine-tuning stage.

In this paper, we focus on the single-identity scenario, and improving the performance of Concat-ID in multiple-identity and multi-subject settings is left for future work. To maximize model performance, we independently train different specialized models for specific tasks. The development of a comprehensive model capable of addressing multiple tasks simultaneously remains a direction for future research.

\section{Ablation study}
\cref{tab:quantitative_abalation} presents the quantitative ablation study of Concat-ID. The pre-training stage achieves the best identity consistency (i.e., ArcSim and CurSim) but has the worst facial editability (i.e., CLIPDist ). However, the cross-video stage significantly improves CLIPDist but degrades ArcSim and CurSim. In the third stage, Concat-ID obtains the second-best results across all metrics, demonstrating that it achieves an optimal balance. These results highlight the superiority of our multi-stage training strategy, which balances the knowledge learned in different stages to achieve optimal performance in the final stage.

Trade-off pairs can naturally enhance the identity consistency of Concat-ID, as they maintain better alignment between reference images and videos compared to cross-video pairs. An interleaved training strategy—alternating between Stage I for improving identity and Stage II for enhancing editability—can also achieve a favorable trade-off, a method similarly adopted in Imagine-yourself~\cite{he2024imagine}. However, our multi-stage training approach achieves an optimal balance just by adding a third stage where we carefully control identity consistency and sample quantity, showing that a simple design can be highly effective.

\begin{table}[tb]
\setlength{\tabcolsep}{5pt}
  \centering
% \resizebox{\linewidth}{!}{
  \begin{tabular}{lcccc}
    \toprule
\multirow{2}[1]{*}{Method}  & \multicolumn{2}{c}{\textbf{Identity consistency}} & \multicolumn{1}{c}{\textbf{Text alignment}} & \multicolumn{1}{c}{\textbf{Facial editability}} \\ 
\cmidrule(lr){2-3} \cmidrule(lr){4-4} \cmidrule(lr){5-5}
 &  ArcSim $\uparrow$ & CurSim $\uparrow$ & ViCLIP $\uparrow$ & CLIPDist $\uparrow$ \\
    \midrule
    Concat-ID (Stage I) &  \bf{0.560} & \bf{0.581} & 0.237 & 0.274   \\
    Concat-ID (Stage II) & 0.185 & 0.200 & \bf{0.248} & \bf{0.434} \\
    Concat-ID (Stage III) & \underline{0.442} & \underline{0.466} & \underline{0.242} & \underline{0.325} \\
    \bottomrule
  \end{tabular}
% }
  \caption{\textbf{Quantitative ablation.} Stage I, Stage II, and Stage III indicate the pre-training stage, cross-video stage, and trade-off stage of Concat-ID, respectively.  The second-best result is underlined. Concat-ID in the third stage demonstrates the optimal balance.}
  \label{tab:quantitative_abalation}
% \vspace{-10pt}
\end{table}

\end{document}